\documentclass[11pt]{article}
\usepackage{nodalida2019}
\usepackage{times}
\usepackage{url}
\usepackage{latexsym}
\usepackage{tikz}
\usepackage{pgfplots}
\usepackage{arydshln}
\usepackage{graphicx}
\usepackage{color, colortbl}
\usepackage{collcell}
\usepackage{hhline}
\usepackage{pgf}
\usepackage{multirow}
\aclfinalcopy 
\usepackage{enumitem}
\setitemize{noitemsep,topsep=2pt,parsep=2pt,partopsep=2pt}

\title{Predicting Prosodic Prominence from Text with Pre-trained Contextualized Word Representations}

\author{
Aarne Talman,\textsuperscript{$\ast$ $\dagger$} Antti Suni,\textsuperscript{$\ast$} Hande Celikkanat,\textsuperscript{$\ast$} Sofoklis Kakouros,\textsuperscript{$\ast$}\\{\bf J\"org Tiedemann\textsuperscript{$\ast$}} and {\bf Martti Vainio\textsuperscript{$\ast$}}\\
\textsuperscript{$\ast$}Department of Digital Humanities, University of Helsinki, Finland\\
\textsuperscript{$\dagger$}Basement AI, Finland \\
\texttt{\{name.surname\}@helsinki.fi}
}

\date{}

\begin{document}
\maketitle

\begin{abstract}
 In this paper we introduce a new natural language processing dataset and benchmark for predicting prosodic prominence from written text. To our knowledge this will be the largest publicly available dataset with prosodic labels. We describe the dataset construction and the resulting benchmark dataset in detail and train a number of different models ranging from feature-based classifiers to neural network systems for the prediction of discretized prosodic prominence. We show that pre-trained contextualized word representations from BERT outperform the other models even with less than 10\% of the training data. Finally we discuss the dataset in light of the results and point to future research and plans for further improving both the dataset and methods of predicting prosodic prominence from text. The dataset and the code for the models are publicly available.
\end{abstract}

\section{Introduction}
Prosodic prominence, i.e., the amount of emphasis that a speaker gives to a word, has been widely studied in phonetics and speech processing. However, the research on text-based natural language processing (NLP) methods for predicting prosodic prominence is somewhat limited. Even in the text-to-speech synthesis domain, with many recent methodological advances, work on symbolic prosody prediction has lagged behind. We believe that this is mainly due to the lack of suitable datasets. Existing, publicly available annotated speech corpora, are very small by current standards.

In this paper we introduce a new NLP dataset and benchmark for predicting prosodic prominence from text which is based on the recently published LibriTTS corpus \cite{zen2019libritts}, containing automatically generated prosodic prominence labels for over 260 hours or 2.8 million words of English audio books, read by 1230 different speakers. To our knowledge this will be the largest publicly available dataset with prosodic annotations. We first give some background about prosodic prominence and related research in Section \ref{sec:background}. We then describe the dataset construction and annotation method in Section \ref{sec:dataset}.

Prosody prediction can be turned into a sequence labeling task by giving each word in a text a discrete prominence value based on the amount of emphasis the speaker gives to the word when reading the text. In Section \ref{sec:experiments} we explain the experiments and the experimental results using a number of different sequence labeling approaches and show that pre-trained contextualized word representations from BERT \cite{devlin2019bert} outperform our other baselines even with less than 10\% of the training data.
Although BERT has been previously applied in various sequence labeling tasks, like named entity recognition \cite{devlin2019bert}, to the best of our knowledge, this is the first application of BERT in the task of predicting prosodic prominence. We analyse the results in Section \ref{sec:analysis}, comparing BERT to a bidirectional long short-term memory (BiLSTM) model and looking at the types of errors made by these selected models. We find that BERT outperforms the BiLSTM model across all the labels. 

Finally in Section \ref{sec:discussion} we discuss the methods in light of the experimental results and highlight areas that are known to negatively impact the results. We also discuss the relevance of pre-training for the task of predicting prosodic prominence.
We conclude by pointing to future research both in developing better methods for predicting prosodic prominence but also to further improve the quality of the dataset. The dataset and the PyTorch code for the models are available on GitHub: \url{https://github.com/Helsinki-NLP/prosody}.


\begin{figure*}[ht!]
\centering
\includegraphics[width=1\linewidth]{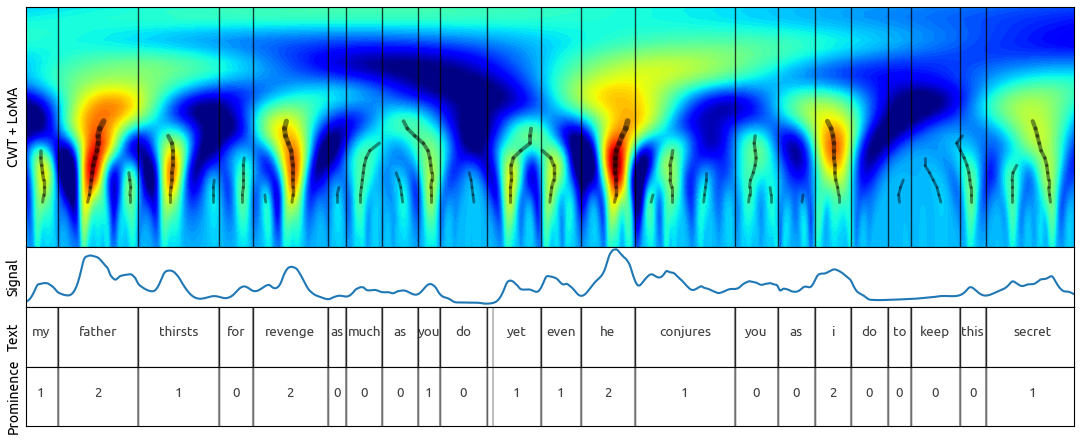}
  \caption{Continuous Wavelet Transform Annotation method.}
    \label{fig:cwt_pic}
\end{figure*}

\section{Background} \label{sec:background}
\subsection{Prosodic Prominence}
Every word and utterance in speech encompasses phonetic and phonological properties that are not resulting from the choice of the underlying lexical items and that encode meaning in addition to that of the individual lexemes. These properties are referred to as prosody and they depend on a variety of factors such as the semantic and syntactic relations between these items, and their rhythmic grouping \cite{wagner2010experimental}. Prosodic variation in speech contributes to a large extend to the perception of natural sounding speech. Prosodic prominence represents one type of prosodic phenomenon that manifests through the subjective impression of emphasis in speech where certain words are interpreted as more salient within their lexical surrounding context \cite{wagner2010experimental,terken2000perception}. 

Due to the inherent difficulty in determining prominence \textemdash ~even for human subjects, see, e.g., \cite{yoon2004intertranscriber} \textemdash ~the development of automatic tools for the annotation of prominent units has been a difficult task. This is exemplified from the large degree of discrepancy observed between human annotators when labeling prominence where the inter-transcriber agreement can vary substantially based on a multitude of factors such as the choice of annotators or annotation method \cite{mo2008naive,yoon2004intertranscriber,kakouros20163pro}.
Similarly, in prominence production, certain degree of freedom in prominence placement and large variability between styles and speakers \cite{yuan2005pitch}, renders the task of prominence prediction from text very difficult compared to most NLP tasks involving text only.

\begin{table*}[ht!]
    \centering
\noindent\begin{tabular}{c|rrr|r|rr}\hline
& & & & \bf non-prominent & \bf prominent &\\
\hline
\bf sets (clean) & \bf speakers & \bf sentences & \bf words & \bf 0  & \bf 1 & \bf 2 \\
\hline
train-100 & 247 &  33,041 & 570,592 & 274,184 & 155,849 & 140,559 \\
train-360 & 904 & 116,262 & 2,076,289 & 1,003,454  & 569,769 & 503,066  \\
dev & 40 & 5,726 & 99,200 & 47,535 & 27,454 & 24,211 \\
test & 39 & 4,821 & 90,063 & 43,234 & 24,543 & 22,286 \\
\hline
total: & 1230 & 159,850 & 2,836,144 & 1,368,407 & 777,615 &  690,122     \\
\hline
\end{tabular}
\caption{Dataset statistics}
    \label{tab:dataset}
\end{table*}

\subsection{Generating Prominence Annotations}
Throughout the literature a number of methods have been proposed for the labeling of prosodic prominence. These methods can be roughly categorized on the basis of the need for training data (manual prosodic annotations) into supervised and unsupervised, but crucially, on the basis of the information they utilize from speech and language to generate their predictions (prominence labels). As prominence perception has been found to correlate with acoustic-phonetic features \cite{lieberman1960some}, with the constituent syntactic structure of an utterance \cite{gregory2004using,wagner2010experimental,bresnan1973sentence}, with the frequency of occurrence of individual lexical items \cite{nenkova2007memorize,jurafsky2001probabilistic}, and with the probabilities of contiguous lexical sequences \cite{jurafsky1996probabilistic}, automatic methods have been developed utilizing these features either in combination or independently \cite{nenkova2007memorize,kakouros2016analyzing,ostendorf1995boston,levow2008automatic}. 

Overall, these features can be largely divided into two categories: (i) \textit{acoustic} (derived from the sound pressure waveform of the speech signal) and (ii) \textit{language} (extracted by studying the form of the language; for instance, semantic or syntactic factors in the language). Both acoustic and language-based features have been shown to provide good overall performance in detecting prominence (in both supervised and unsupervised cases), where, however, the methods utilizing acoustic features seem to provide better performance for the unsupervised detection of prominences in speech \cite{suni2017hierarchical,wang2007acoustic,kakouros20163pro}, with state-of-the-art results reaching high level of accuracy, close to that of the inter-annotator agreement for the data. While the top-down linguistic information is known to correlate with perceptual prominence, in this paper we want to make a clear distinction between data labelling and text-based prediction. Thus, in this work, we utilize purely acoustic prominence annotations of the speech data using the method developed by \citet{suni2017hierarchical} as the prosodic reference.

\subsection{Predicting Prosodic Prominence from Text}
To what extent prosodic prominence can be predicted from textual input only has been a topic of inquiry in linguistics for a long time. In traditional generative phonology \cite{chomsky1968sound}, accent placement was considered to be fully determined by linguistic structure, whereas a seminal work by \citet{bolinger1972accent} emphasized the importance and relevance of the lexical semantic context as well as the speakers' intention, positing that, in general, a mind reading ability may be necessary to determine prominent words in a sentence. As longstanding inquiries hold, the goal of reliably predicting the placement of prominent entities from information automatically derived from textual resources is still ongoing. 

Several efforts have been made towards this direction, especially in text-to-speech (TTS) synthesis research, where generation of appropriate prosody would increase both intelligibility and quality of synthetic speech.
Before the deep learning paradigm shift in NLP, several linguistic features were examined for prominence prediction, including function-content word distinction, part-of-speech class, and information status \cite{hirschberg1993pitch}. Statistical features like unigrams, bigrams, and TF-IDF have also been frequently used \cite{Marsi}.  Later, the accent ratio, or simply the average accent status of a word type in the given corpus, was found to be a stronger predictor than linguistic features in the accent prediction task \cite{nenkova2007memorize}, suggesting that lexical information may be more relevant than linguistic structure for the prominence prediction task.

Recently, continuous representations of words have become commonplace in prosody prediction for TTS, though the symbolic level is often omitted and pitch and duration are predicted directly using lexical embeddings \cite{watts-2012}.
Yet, closely related to the proposed method, \cite{rendel2016using} experimented with various lexical embeddings as an input to a Bi-directional LSTM model, predicting binary prominence labels. Training on a proprietary, manually annotated single speaker corpus of 3730 sentences, they achieved an F-score of 0.71 with Word2Vec \cite{mikolov2014word2vec} embeddings, with a clear improvement over traditional linguistic features.


\begin{table*}[ht!]
    \centering
    \noindent\begin{tabular}{l|cccccccccc}
    \hline
\bf Token & Tell & me & you & rascal & , & where & is & the & pig & ? \\
\hline
\bf Discrete label & 2 & 0 & 0 & 0 & NA & 2 & 0 & 0 & 1 & NA\\
\cdashline{1-11}
\bf Real-valued label & 1.473 & 0.333 & 0.003 & 0.167 & NA & 2.160 & 0.006 & 0.037 & 0.719 & NA\\
\hline
\end{tabular}
\caption{Example sentence with the annotation from the dataset. Discrete prominence values were used in the experiments of this paper. The real-valued labels are used for generation of the discrete labels, however, they could also be used directly for prominence prediction.}
    \label{tab:example_data}

\end{table*}

\section{Dataset} \label{sec:dataset}
We introduce, automatically generated, high quality prosodic annotations for the recently published LibriTTS corpus \cite{zen2019libritts}. The LibriTTS corpus is a cleaned subset of LibriSpeech corpus \cite{panayotov2015librispeech}, derived from English audiobooks of the LibriVox project.\footnote{\url{https://librivox.org}} 
We selected the `clean' subsets of LibriTTS for annotation, comprising of 262.5 hours of read speech from 1230 speakers. The transcribed sentences were aligned with the Montreal forced aligner \cite{mcauliffe2017montreal}, using a pronunciation lexicon and acoustic models trained on the LibriSpeech dataset. The aligned sentences were then prosodically annotated with word-level acoustic prominence labels. For the annotation, we used the Wavelet Prosody Analyzer toolkit\footnote{\url{https://github.com/asuni/wavelet_prosody_toolkit}}, which implements the method described in \cite{suni2017hierarchical}. Briefly, the method consists of 1) the extraction of pitch and energy signals from the speech data and duration from the word level alignments, 2) filling the unvoiced gaps in extracted signals by interpolation followed by smoothing and normalizing, 3) combining the normalized signals by summing or multiplication, and 4) performing a continuous wavelet transform (CWT) on the composite signal and extracting continuous prominence values as lines of maximum amplitude across wavelet scales (see Figure \ref{fig:cwt_pic}).  Essentially, the method assumes that the louder, the longer, and the higher, the more prominent. On top of this, the wavelet transform provides multi-resolution contextual information; the more the word stands out from its environment in various time scales, the more prominent the word is perceived.

For the current study, continuous prominence values were discretized to two (non-prominent, prominent) or three (non prominent, somewhat prominent, very prominent) classes. The binary case is closely related to the pitch accent detection task, aiming
for results comparable with the majority of the literature on the topic. The weights in constructing the composite signal and discretization thresholds were adjusted based on The Boston University radio news corpus \cite{ostendorf1995boston}, containing manually annotated pitch accent labels. This corpus is often used in the evaluation of pitch accent annotation and prediction quality, with the current annotation method yielding state-of-the-art accuracy in word level acoustic-based accent detection, 85.3\%, using weights 1.0, 0.5 and 1.0 for F0, energy and duration respectively, and using multiplication of these features in signal composition. For three-way discretization, the non-prominent / prominent cut-off was maintained and the prominent class was split to two classes of roughly equal size. Statistics of the resulting dataset are described in table \ref{tab:dataset}. The full dataset is available for download here: \url{https://github.com/Helsinki-NLP/prosody}. Although not discussed in this paper, the described acoustic annotation and text-based prediction methods can be applied to prosodic boundaries too, and the boundary labels will be included in the dataset at a later stage.

\section{Experiments} \label{sec:experiments}
In this section we describe the experimental setup and the results from our experiments in predicting discrete prosodic prominence labels from text using the corpus described above.

\subsection{Experimental Setup}

We performed experiments with the following models:
\begin{itemize}
    \item BERT-base uncased \cite{devlin2019bert}
    \item 3-layer 600D Bidirectional Long Short-Term Memory (BiLSTM) \cite{Hochreiter:1997}
    \item Minitagger (SVM) \cite{stratos-collins-2015-simple} + GloVe \cite{pennington2014glove}
    \item MarMoT (CRF) \cite{mueller-etal-2013-efficient}
    \item Majority class per word
\end{itemize}

The models were selected so that they cover a wide variety of different architectures from feature-based statistical approaches to neural networks and pre-trained language models. The models are described in more detail below.

We use the Huggingface PyTorch implementation of BERT available in the \texttt{pytorch\_transformers} library,\footnote{\url{https://github.com/huggingface/pytorch-transformers}} which we further fine-tune during training. We take the last hidden layer of BERT and train a single fully-connected classifier layer on top of it, mapping the representation of each word to the labels.
For our experiments we use the smaller BERT-base model using the uncased alternative. We use a batch size of 32 and fine-tune the model for 2 epochs.

For BiLSTM we use pre-trained 300D GloVe 840B word embeddings \cite{pennington2014glove}. The initial word embeddings are fine-tuned during training. As with BERT, we add one fully-connected classifier layer on top of the BiLSTM, mapping the representation of each word to the labels. 
We use a dropout of 0.2 between the layers of the BiLSTM. We use a batch size of 64 and train the model for 5 epochs.

For the SVM we use Minitagger\footnote{\url{https://github.com/karlstratos/minitagger}} implementation by \citet{stratos-collins-2015-simple} using each dimension of the pre-trained 300D GloVe 840B word embeddings as features, with context-size 1, i.e. including the previous and the next word in the context.

For the conditional random field (CRF) model we use MarMot\footnote{\url{http://cistern.cis.lmu.de/marmot/}} by \citet{mueller-etal-2013-efficient} with the default configuration. The model applies standard feature templates that are used for part-of-speech tagging such as surrounding words as well as suffix and prefix features. We did not optimize the feature model nor any of the other hyper-parameters.

All systems except the Minitagger and CRF are our implementations using PyTorch and are made available on GitHub: \url{https://github.com/Helsinki-NLP/prosody}.

For the experiments we used the larger train-360 training set. We report both 2-way and 3-way classification results. In the 2-way classification task we take the three prominence labels and merge labels 1 and 2 into a single prominent class.

\subsection{Results}

\begin{table*}[ht!]
    \centering
    \begin{tabular}{l c c}
    \hline
    \bf Model & \bf Test accuracy (2-way) & \bf Test accuracy (3-way) \\
    \hline
        BERT-base & \bf83.2\% & \bf68.6\%\\
        3-layer BiLSTM & 82.1\% & 66.4\% \\
        CRF & 81.8\% & 66.4\%\\
        SVM+GloVe & 80.8\% & 65.4\% \\
        Majority class per word & 80.2\% & 62.4\% \\
        Majority class & 52.0\% & 48.0\% \\ 
        Random & 49.0\% & 39.5\% \\
    \hline
    \end{tabular}
    \caption{Experimental results (\%) for the 2 and 3-way classification tasks.}
    \label{tab:results}
\end{table*}

All models reach over 80\% in the 2-way classification task while 3-way classification accuracy stays below 70\% for all of them. The BERT-based model gets the highest accuracy of 83.2\% and 68.6\% in the 2-way and 3-way classification tasks, respectively, demonstrating the value of a pre-trained language model in this task. The 3-layer BiLSTM achieves 82.1\% in the 2-way classification and 66.4\% in the 3-way classification task.

The traditional feature-based classifiers perform slightly below the neural network models, with the CRF obtaining 81.8\% and 66.4\% for the two classification tasks, respectively. The Minitagger SVM model's test accuracies are slightly lower than the CRF's with 80.8\% and 65.4\% test accuracies. Finally taking a simple majority class per word gives 80.2\% for the 2-way classification task and 62.4\% for the 3-way classification task. The results are listed in Table \ref{tab:results}. The fairly low results across the board highlight the difficulty of the task of predicting prosodic prominence from text. 

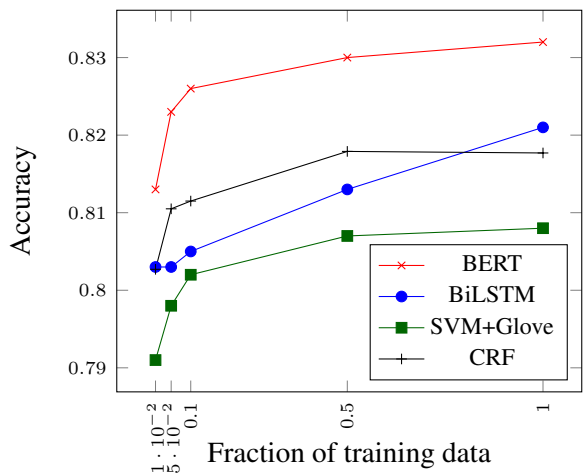
\begin{figure}
\centering
\begin{tikzpicture}
	\begin{axis}[
	    width=\columnwidth,
	    legend style={font=\small},
		xlabel=Fraction of training data,
		ylabel=Accuracy,
		legend pos=south east,
		xtick={0.01, 0.05, 0.1, 0.5,1.0},
		xticklabel style={rotate = 90, font=\scriptsize},
		yticklabel style={font=\scriptsize}
		]
	\addplot[color=red,mark=x, mark options={fill=white}] table[x=Data,y=Accuracy] {
Data    Accuracy
1       0.832
0.5     0.830
0.1     0.826
0.05    0.823
0.01    0.813
	};
		\addplot[color=blue,mark=*] table[x=Data,y=Accuracy] {
Data	Accuracy
1       0.821
0.5     0.813
0.1     0.805
0.05    0.803
0.01    0.803
	};
		\addplot[color=black!60!green,mark=square*] table[x=Data,y=Accuracy] {
Data	Accuracy
1       0.808
0.5     0.807
0.1     0.802
0.05    0.798
0.01    0.791
	};
		\addplot[color=black,mark=+] table[x=Data,y=Accuracy] {
Data	Accuracy
1       0.8177
0.5     0.8179
0.1     0.8115
0.05    0.8105
0.01    0.8027
	};
	\legend{BERT, BiLSTM, SVM+Glove, CRF}
	\end{axis}
\end{tikzpicture}
\caption{Test accuracy with different size subsets of the training data for the 2-way classification task. }
\label{fig:learning_curve_2-way}
\end{figure}

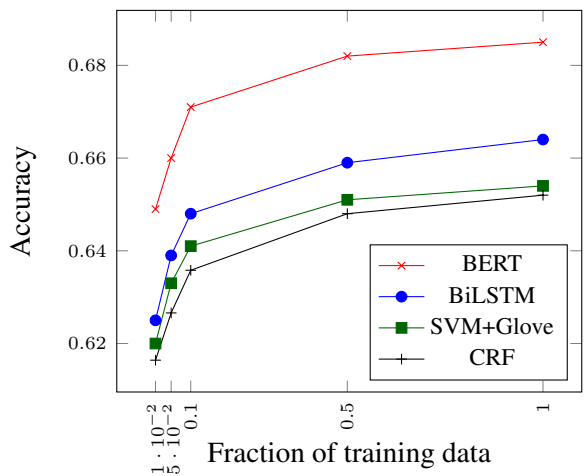
\begin{figure}
\centering
\begin{tikzpicture}
	\begin{axis}[
	    width=\columnwidth,
	    legend style={font=\small},
		xlabel=Fraction of training data,
		ylabel=Accuracy,
		legend pos=south east,
		xtick={0.01, 0.05, 0.1, 0.5,1.0},
		xticklabel style={rotate = 90, font=\scriptsize},
		yticklabel style={font=\scriptsize},
		]
	\addplot[color=red,mark=x, mark options={fill=white}] table[x=Data,y=Accuracy] {
Data    Accuracy
0.01    0.649
0.05    0.660
0.1     0.671
0.5     0.682
1       0.685
	};
	\addplot[color=blue,mark=*] table[x=Data,y=Accuracy] {
Data	Accuracy
0.01	0.625
0.05    0.639
0.1     0.648
0.5     0.659
1       0.664
	};
		\addplot[color=black!60!green,mark=square*] table[x=Data,y=Accuracy] {
Data	Accuracy
0.01	0.620
0.05    0.633
0.1     0.641
0.5     0.651
1       0.654
	};
		\addplot[color=black,mark=+] table[x=Data,y=Accuracy] {
Data	Accuracy
1       0.6520
0.5     0.6480
0.1     0.6358
0.05    0.6266
0.01    0.6164
	};
	\legend{BERT, BiLSTM, SVM+Glove, CRF}
	\end{axis}
\end{tikzpicture}
\caption{Test accuracy with different size subsets of the training data for the 3-way classification task. }
\label{fig:learning_curve}
\end{figure}

To better understand how much training data is needed in the two classification tasks, we trained selected models with different size subsets of the train-360 training data. The selected subsets were: 1\%, 5\%, 10\%, 50\% and 100\% of the training examples (token-label pairs). Figures \ref{fig:learning_curve_2-way} and \ref{fig:learning_curve} contain the learning curves for the 2-way and 3-way classification tasks, for all the models except for the majority and random baselines.

For all models and for both of the classification tasks we notice that they achieve quite high test accuracy already with a very small number of training examples. For most of the models the biggest improvement in performance is achieved when moving from 1\% of the training examples to 5\%. All models have reached close to their full predictive capacity with only 10\% of the training examples. For example, BERT achieves 2-way classification test accuracy of 82.6\% with 10\% of the training data, which is only -0.6\% points lower than the accuracy with the full training set. In the 3-way classification task 10\% of the training data gives 67.1\% for BERT, which is -1.7\% points below the accuracy with the full training set.

Interestingly, in the 2-way classification task the BiLSTM model shows a slightly different learning curve, having already quite a high performance with just 1\% of the training data, but then making no improvement between 1\% and 5\%. However, between 5\% and 100\% the BiLSTM model improvement is almost linear. 

As the proposed dataset has been automatically generated as described in Section \ref{sec:dataset}, we also tested the best two models, BERT and BiLSTM, with a manually annotated test set from The Boston University radio news corpus \cite{ostendorf1995boston}. For this experiment we trained the models using the train-360 training set (as above) replacing only the test set. The results of this experiment are shown in Table \ref{tab:results_boston}. The good results\footnotemark ~from this experiment provide further support for the quality of the new dataset. Notice also that the difference between BERT and BiLSTM is much bigger with this test set (+3.9\% compared to +1.1\%). This difference could be due to the genre difference between the two test sets, with the Boston University news corpus being more contemporary compared to the source for our proposed dataset (pre-1923 books). This point will be further discussed in Section \ref{sec:discussion}.
\footnotetext{Better results have been reported on Boston dataset using lexical features, but there are methodological concerns related to cross-validation training and speakers reading the same text, see discussion on \cite{Rosenberg2009}.}

\begin{table}[h!]
    \centering
    \begin{tabular}{lcc}
    \hline
    \bf Model & \bf vs expert & \bf vs acoustic\\
    \hline
        BERT-base & \bf 82.9\% & \bf 82.1\% \\
        3-layer BiLSTM & 79.0\% & 79.3\% \\
    \hline
    \end{tabular}
    \caption{Test accuracies (\%) for the Boston University radio news corpus (2-way classification). expert = expert annotated perceptual prominence labels, acoustic = our acoustic prominence labels }
    \label{tab:results_boston}
\end{table}

\section{Analysis}\label{sec:analysis}

The experimental results show that although predicting prosodic prominence is a fairly difficult task, pre-trained contextualized word representations clearly help, as can be seen from the results for BERT. The difference between BERT and the other models is clear if we compare the other models with BERT fine-tuned with a small fraction of the training data. In fact, BERT already outperforms the other models with just 5\% of the training examples in the 2-way classification case and with 10\% of the training data in the 3-way classification case. This can be seen as an indication that BERT has acquired implicit semantic or syntactic information during pre-training that is useful in the task of predicting prosodic prominence.

To gain a better understanding of the types of predictive errors BERT makes, we look at the confusion matrices for the two classification tasks and compare those with the confusion matrices for the BiLSTM.

\begin{figure}[h!]
    \centering
    \includegraphics[width=\linewidth]{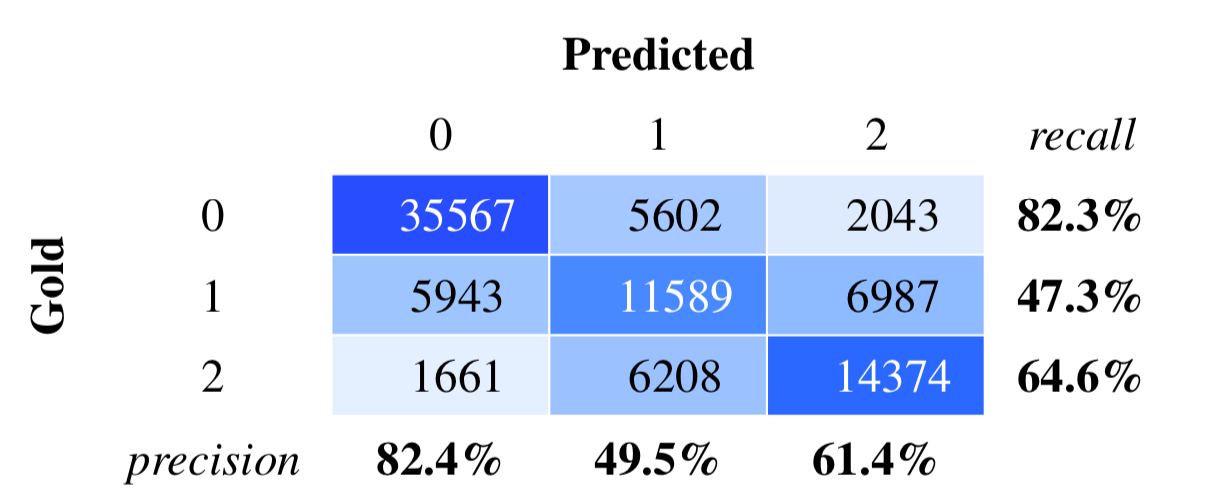}
    \caption{3-way classification task confusion matrix for BERT.}
    \label{figure:BERTconf_3way}
\end{figure}

\begin{figure}[h!]
    \centering
    \includegraphics[width=\linewidth]{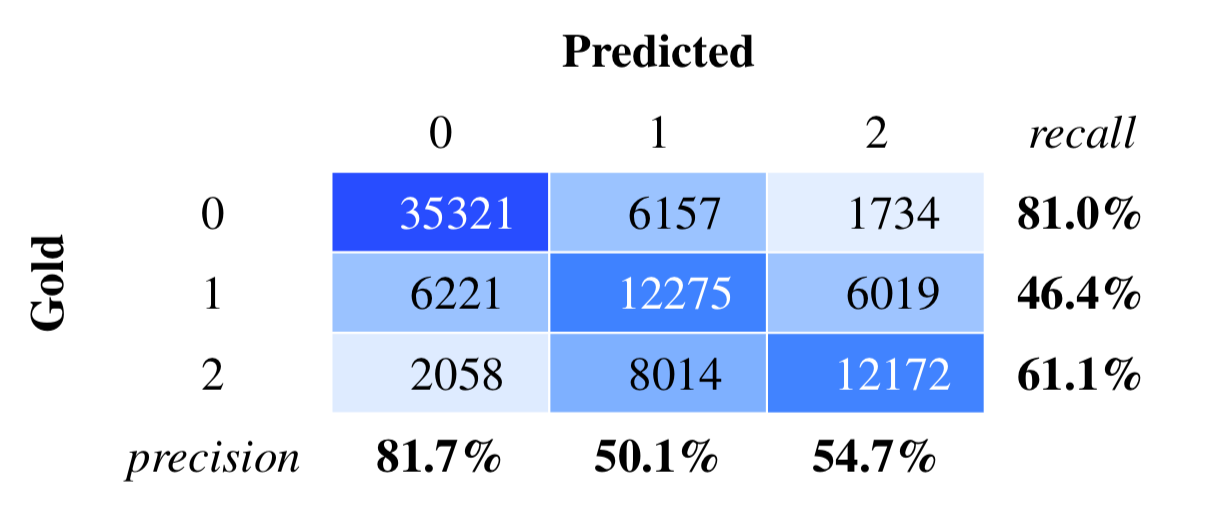}
    \caption{3-way classification task confusion matrix for BiLSTM.}
    \label{figure:BiLSTMconf_3way}
\end{figure}

The 3-way classification confusion matrices are more informative as they allow comparison of the two models with respect to the predicted label in cases of error. Figure \ref{figure:BERTconf_3way} contains the 3-way classification confusion matrix for BERT and Figure \ref{figure:BiLSTMconf_3way} for the BiLSTM model. 


In the 3-way classification task, when the gold label is 0 (non prominent) BERT makes more errors with prediction being 2 (very prominent) compared to the BiLSTM model. However, when the gold label is 2 (very prominent) BiLSTM makes more predictions with 0 (non prominent) compared to BERT. In general for 0 labels BERT seems to have higher precision and BiLSTM better recall, whereas for label 2 BERT has clearly higher recall and precision. Both models have low precision and recall for the less distinctive prominence (label 1). It seems that the clearest difference between the two models is in their ability to predict high prominence (label 2). 

\begin{figure}[h!]
    \centering
    \includegraphics[width=\linewidth]{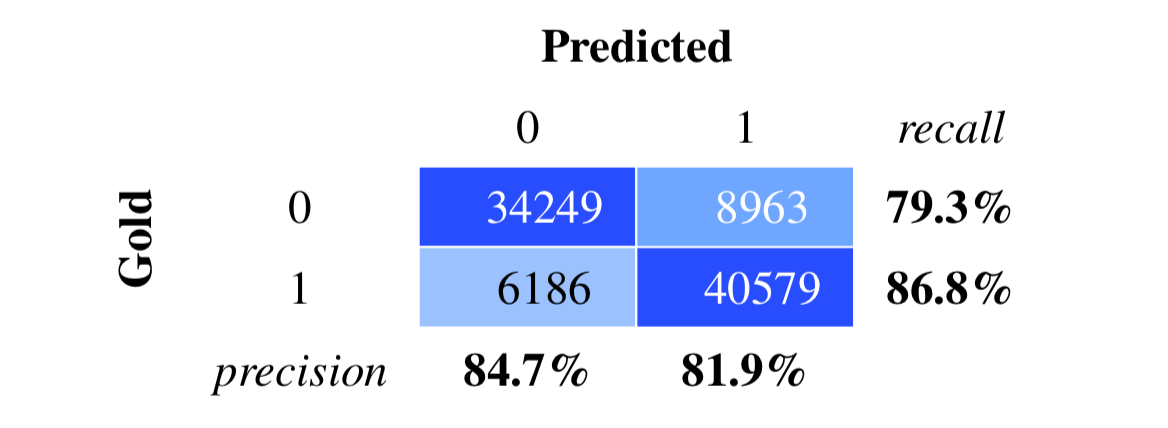}
    \caption{2-way classification task confusion matrix for BERT.}
    \label{figure:BERTconf_2way}
\end{figure}

\begin{figure}[h!]
    \centering
    \includegraphics[width=\linewidth]{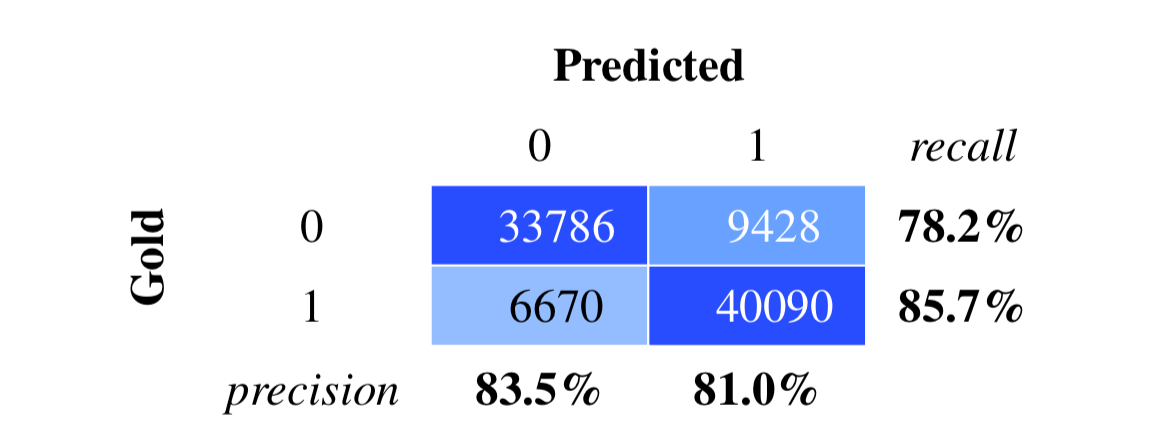}
    \caption{2-way classification task confusion matrix for BiLSTM.}
    \label{figure:BiLSTMconf_2way}
\end{figure}

We also provide the confusion matrices for the 2-way classification task for the two models. Figure \ref{figure:BERTconf_2way} contains the 2-way classification confusion matrix for BERT and Figure \ref{figure:BiLSTMconf_2way} for the BiLSTM model. Here BERT has slightly higher precision and recall across both of the labels.


\begin{table*}[h!]

\begingroup
\setlength{\tabcolsep}{0.2\tabcolsep}%
\noindent\begin{tabular}{*{20}{l}}\\
REF: &  \textbf{One} & way & {\tiny led} & {\tiny to} & {\tiny the} & left & {\tiny and} & {\tiny the} & {\tiny other} & {\tiny to} & {\tiny the} & right & \textbf{straight} & {\tiny up} & {\tiny the} & mountain & . &  \\
BERT:\quad~ &  \textbf{One} & {\tiny way} & led & {\tiny to} & {\tiny the} & \textbf{left} & {\tiny and} & {\tiny the} & \textbf{other} & {\tiny to} & {\tiny the} & right & \textbf{straight} & {\tiny up} & {\tiny the} & mountain & . &  \\
\vspace{-8.5mm}
\end{tabular}
\\
\noindent\begin{tabular}{*{20}{l}}\\
REF: &  In & {\tiny the} & \textbf{next} & moment & {\tiny he} & {\tiny was} & \textbf{concealed} & {\tiny by} & {\tiny the} & leaves & . &  \\
BERT:\quad~ &  {\tiny In} & {\tiny the} & \textbf{next} & moment & {\tiny he} & {\tiny was} & \textbf{concealed} & {\tiny by} & {\tiny the} & leaves & . &  \\
\vspace{-8.5mm}
\end{tabular}
\\
\noindent\begin{tabular}{*{20}{l}}\\
REF: &  {\tiny I} & \textbf{had} & {\tiny to} & {\tiny read} & {\tiny it} & {\tiny over} & \textbf{carefully} & , & {\tiny as} & {\tiny the} & \textbf{text} & {\tiny must} & {\tiny be} & {\tiny absolutely} & correct & . &  \\
BERT:\quad~ &  {\tiny I} & \textbf{had} & {\tiny to} & read & {\tiny it} & over & \textbf{carefully} & , & as & {\tiny the} & \textbf{text} & must & {\tiny be} & absolutely & \textbf{correct} & . &  \\
\vspace{-8.5mm}
\end{tabular}
\\
\noindent\begin{tabular}{*{20}{l}}\\
REF: &  \textbf{Where} & were & {\tiny you} & {\tiny when} & {\tiny you} & \textbf{began} & {\tiny to} & \textbf{feel} & bad & ? &  \\
BERT:\quad~ &  Where & \textbf{were} & {\tiny you} & {\tiny when} & {\tiny you} & began & {\tiny to} & feel & \textbf{bad} & ? &  \\
\vspace{-8.5mm}
\end{tabular}
\\
\noindent\begin{tabular}{*{20}{l}}\\
REF: &  {\tiny He} & {\tiny is} & taller & than & {\tiny the} & {\tiny Indian} & , & \textbf{not} & {\tiny so} & \textbf{tall} & {\tiny as} & \textbf{Gilchrist} & . &  \\
BERT:\quad~ &  {\tiny He} & {\tiny is} & \textbf{taller} & {\tiny than} & {\tiny the} & Indian & , & \textbf{not} & so & tall & {\tiny as} & \textbf{Gilchrist} & . &  \\
\vspace{-8.5mm}
\end{tabular}
\\
\endgroup
\caption{Typical 3-way prominence predictions of BERT compared to reference labels. 
}
\label{predictions}
\end{table*}

\section{Discussion}\label{sec:discussion}
We have shown above that prosodic prominence can reasonably well be predicted from text using different sequence-labelling approaches and models. However, the reported performance is still quite low, even for state-of-the-art systems based on large pre-trained language models such as BERT. We list a number of reasons for these shortcomings below and discuss their impact and potential mitigation. 

Although the annotation method has been shown to be quite robust, errors in automatic alignment, signal processing, and quantization introduce noise to the labels. This noise might not be detrimental to the training due to dataset size, but the test results are affected. To measure the size of this effect, manual correction of a part of the test set could be beneficial.

It is well known that different speakers have different accents, varying reading proficiency, and reading tempo, which all impact the consistency of the labeling as the source speech data contains in total samples from over 1200 different speakers. Given that inter-speaker agreement on pitch accent placement is somewhere between 80 and 90\% \cite{yuan2005pitch}, we cannot expect large improvements without speaker-specific modelling.

The source speech data contains multitude of genres ranging from non-fiction to metric poems with fixed prominence patterns and children's stories with high proportion of words emphasized. The difference in genres could impact the test results. Moreover, the books included in the source speech data are all from pre-1923, whereas BERT and GloVe are pre-trained with contemporary texts. We expect that the difference between BERT and other models would be higher with a dataset drawn from a more contemporary source. As noted in Section \ref{tab:results}, the difference between BERT and BiLSTM is much bigger with the The Boston University radio news corpus test set (+3.9\% compared to +1.1\% with our test set). This could be due to the genre, with The Boston University radio news corpus being derived from a more contemporary source.

Overall, our results for BERT highlight the importance of pre-training of the word representations. As we noticed, already with as little as 10\% of the training data, BERT outperforms the other models when they are trained on the entire training set. This suggests that BERT has implicitly learned syntactic or semantic information relevant for the prosody prediction task. Our results are in line with the earlier results by \citet{stehwien2018embeddings} and \citet{rendel2016using} who showed that pre-trained word embeddings improve model performance in the prominence prediction task. Table \ref{predictions} lists five randomly selected examples from the test set and shows the prominence predictions by BERT compared to the reference annotation. These examples indicate that even if the overall accuracy of the model is not high, the predictions still look plausible in isolation.

Finally, the classifiers in this paper are trained on single sentences, losing any discourse-level information and relations to surrounding context. Increasing the context to contain, e.g., also previous sentences could improve the results.

\section{Conclusion}
In this paper we have introduced a new NLP dataset and benchmark for predicting prosodic prominence from text, which to our knowledge is the largest publicly available dataset with prosodic labels. We described the dataset creation and the resulting benchmark and showed that various sequence labeling methods can be applied to the task of predicting prosodic prominence using the dataset.

Our experimental results show that BERT outperforms the other models with just up to 10\% of the training data, highlighting the effectiveness of pre-training for the task. It also highlights that the implicit syntactic or semantic features BERT has learned during pre-training are relevant for the specific task of predicting prosodic prominence.

We also discussed a number of limitations of the automatic annotation system, as well as our current models. Based on this discussion, and more broadly, on the findings of this paper, we want to focus our future research activities in two fronts. Firstly, we will further develop the dataset annotation pipeline, improving the quality of prominence annotation and adding prosodic boundary labels. Secondly, we will further develop methods and models for improved prediction of prosodic prominence. In particular, as our results have shown that pre-training helps in the task, fine-tuning BERT with data involving features that are known to impact prosodic prominence (like part-of-speech tagged data) before training on the prosody dataset could help to improve the model performance. Furthermore, we will look at speaker-aware models, genre adaptation, and models for increased context. And, finally, our ultimate goal is to incorporate these methods into the development of a state-of-the-art text-to-speech synthesizer.

\section*{Acknowledgments}
\vspace{1ex}
\noindent
\begin{minipage}{0.1\linewidth}
   \raisebox{-0.2\height}{\includegraphics[trim =32mm 55mm 30mm 5mm, clip, scale=0.2]{erc.ai}}
\end{minipage}
\hspace{0.01\linewidth}
\begin{minipage}{0.70\linewidth}
Talman, Celikkanat and Tiedemann are supported by the FoTran project, funded by the European Research Council (ERC) under the European Union’s Horizon 2020 research and innovation programme (grant agreement no.~771113). 

 \vspace{1ex}
\end{minipage}
\hspace{0.01\linewidth}
\begin{minipage}{0.05\linewidth}
 \vspace{0.05cm}
\raisebox{-0.25\height}{\includegraphics[trim =0mm 5mm 5mm 2mm,clip,scale=0.078]{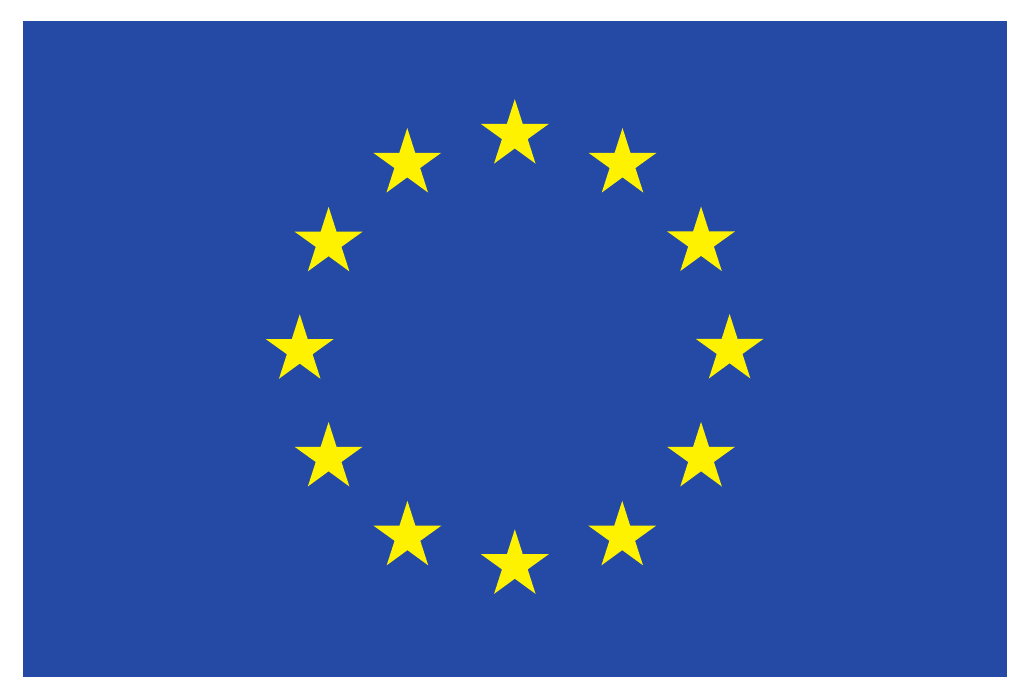}}
\vspace{0.05cm}

\end{minipage}

We also gratefully acknowledges the support of the Academy of Finland through projects no.~314062 from the ICT 2023 call on Computation, Machine Learning and Artificial Intelligence, no.~1293348 from the call on Digital Humanities, and an Academy Fellowship project no.~309575.

\bibliographystyle{acl_natbib}
\bibliography{references}

\end{document}